%% file: main.tex
\documentclass[10pt, a4paper]{article}
\usepackage{lrec}
\usepackage{graphicx}
\usepackage{tabularx}
\usepackage{booktabs}
\usepackage{soul}
\usepackage{titlesec}
\titleformat{\section}{\normalfont\large\bf\center}{\thesection.}{1em}{}
\titleformat{\subsection}{\normalfont\SmallTitleFont\bf\raggedright}{\thesubsection.}{1em}{}
\titleformat{\subsubsection}{\normalfont\normalsize\bf\raggedright}{\thesubsubsection.}{1em}{}
\renewcommand\thesection{\arabic{section}}
\renewcommand\thesubsection{\thesection.\arabic{subsection}}
\renewcommand\thesubsubsection{\thesubsection.\arabic{subsubsection}}
\usepackage[latin1]{inputenc}
\usepackage[hidelinks]{hyperref}
\usepackage{xstring}
\usepackage[dvipsnames]{xcolor} 
\usepackage{gb4e} 
\noautomath % always with gb4e

\title{A Dataset of German Legal Documents for Named Entity Recognition}

\name{Elena Leitner, Georg Rehm, Juli\'{a}n Moreno-Schneider}

\address{DFKI GmbH, Alt-Moabit 91c, 10559 Berlin, Germany\\
         \{firstname.lastname\}@dfki.de\\}

\abstract{We describe a dataset developed for Named Entity Recognition in German federal court decisions. It consists of approx.~67,000 sentences with over 2 million tokens. The resource contains 54,000 manually annotated entities, mapped to 19 fine-grained semantic classes: \emph{person}, \emph{judge}, \emph{lawyer}, \emph{country}, \emph{city}, \emph{street}, \emph{landscape}, \emph{organization}, \emph{company}, \emph{institution}, \emph{court}, \emph{brand}, \emph{law}, \emph{ordinance}, \emph{European legal norm}, \emph{regulation}, \emph{contract}, \emph{court decision}, and \emph{legal literature}. The legal documents were, furthermore, automatically annotated with more than 35,000 TimeML-based time expressions. The dataset, which is available under a CC-BY~4.0 license in the CoNNL-2002 format, was developed for training an NER service for German legal documents in the EU project Lynx.
\\ \newline }
\begin{document}

\maketitleabstract
\section{Introduction and Motivation}

Just like any other field, the legal domain is facing multiple challenges in the era of digitisation. Document collections are growing at an enormous pace and their complete and deep analysis can only be tackled with the help of assisting technologies. This is where content curation technologies based on text analytics come in \newcite{rehm2016j}. Such domain-specific semantic technologies enable the fast and efficient automated processing of heterogeneous document collections, extracting important information units and metadata such as, among others, named entities, numeric expressions, concepts and topics, time expressions, and text structure. One of the fundamental processing tasks is the identification and categorisation of named entities (Named Entity Recognition, NER). Typically, NER is focused upon the identification of semantic categories such as \emph{person}, \emph{location} and \emph{organization} but, especially in domain-specific applications, other typologies have been developed that correspond to task-, language- or domain-specific needs. With regard to the legal domain, the lack of freely available datasets has been a stumbling block for text analytics research. German newspaper datasets from CoNNL 2003 \cite{tjong2003introduction} or GermEval 2014 \cite{benikova2014nosta} are simply not suitable in terms of domain, text type or semantic categories covered.

The work described in this paper was carried out under the umbrella of the project \emph{Lynx: Building the Legal Knowledge Graph for Smart Compliance Services in Multilingual Europe}, a three-year EU-funded project that started in December 2017 \cite{montiel2017}.\footnote{\url{http://www.lynx-project.eu}} Its objective is the creation of a legal knowledge graph that contains different types of legal and regulatory data \cite{rehm2018g,rehm2018f,morenoschneider2020j}. Lynx aims to help European companies, especially SMEs, that want to become active in new European countries and markets. The project offers compliance-related services that are currently tested and validated in three use cases (UC):
(i) UC1 aims to analyse contracts, enriching them with domain-specific semantic information (document structure, entities, temporal expressions, claims, summaries, etc.);
(ii) UC2 focuses on compliance services related to geothermal energy operations, where Lynx supports the understanding of regulatory regimes, including norms and standards; (iii) UC3 is a compliance solution in the domain of labour law, where legal provisions, case law, and expert literature are interlinked, analysed, and compared to define legal strategies for legal practice. The Lynx services are developed for several European languages including English, Spanish, and -- relevant for this paper -- German \cite{rehm2019c}.

Documents in the legal domain contain multiple references to named entities, especially domain-specific named entities, i.\,e., jurisdictions, legal institutions, etc. Legal documents are unique and differ greatly from newspaper texts. On the one hand, the occurrence of general-domain named entities is relatively rare. On the other hand, in concrete applications, crucial domain-specific entities need to be identified in a reliable way, such as designations of legal norms and references to other legal documents (laws, ordinances, regulations, decisions, etc.). However, most NER solutions operate in the general or news domain, which makes them inapplicable to the analysis of legal documents \cite{rehm2017a,rehm2017h}. Accordingly, there is a great need for an NER-annotated dataset consisting of legal documents, including the corresponding development of a typology of semantic concepts and uniform annotation guidelines. In this paper, we describe the development of a dataset of legal documents, which includes (i) named entities and (ii) temporal expressions.

The remainder of this article is structured as follows. First, Section~\ref{sec:relatedwork} gives a brief overview of related work. Section~\ref{sec:annotationguidelines} describes, in detail, the rationale behind the annotation of the dataset including the different semantic classes annotated. Section~\ref{sec:dataset} describes several characteristics of the dataset, followed by a short evaluation (Section~\ref{sec:evaluation}) and conclusions as well as future work (Section~\ref{sec:conclusion}).

\section{Related Work}
\label{sec:relatedwork}

Until now, NER has not received a lot of attention in the legal domain, developed approaches are fragmented and inconsistent with regard to their respective methods, datasets and typologies used. Among the related work, there is no agreement regarding the selection of relevant semantic categories from the legal domain. In addition, corpora or datasets of legal documents with annotated named entities do not appear to exist, which is, obviously, a stumbling block for the development of data-driven NER classifiers. 

\newcite{dozier2010named} describe five classes for which taggers are developed based on dictionary lookup, pattern-based rules, and statistical models. These are \emph{jurisdiction} (a geographic area with legal authority), \emph{court}, \emph{title} (of a document), \emph{doctype} (category of a document), and \emph{judge}. The taggers were tested with documents such as US case law, depositions, pleadings etc. \newcite{cardellino2017low} develop an ontology of legal concepts, making use of NERC (6 classes), LKIF (69 classes) and YAGO (358 classes). On the NERC level, entities were divided in \emph{abstraction}, \emph{act}, \emph{document}, \emph{organization}, \emph{person}, and non-entity. With regard to LKIF, \emph{company}, \emph{corporation}, \emph{contract}, \emph{statute} etc.~are used. Unfortunately, the authors do not provide any details regarding the questions how the entities were categorised or if there is any correlations between the different levels. They work with Wikipedia articles and decisions of the European Court of Human Rights. \newcite{glaser2017named} use GermaNER \cite{benikova2015c} and DBpedia Spotlight \cite{mendes2011dbpedia,isem2013daiber} for the recognition of \emph{person}, \emph{location} and \emph{organization} entities. References are identified based on the rules described by \newcite{landthaler2016unveiling}. The authors created an evaluation dataset of 20 court decisions.

\section{Annotation of the Dataset}
\label{sec:annotationguidelines}

In the following, we describe the rationale behind the annotation of the dataset including the definition of the various semantic classes and the annotation guidelines.

\subsection{Named Entities vs.~Legal Entities}

\paragraph{Named Entity} An entity is an object or set of objects in the real world and can be referenced in a text with a proper name, noun or pronoun \cite{ace_guidelines}. The examples (\ref{name}--\ref{pronoun}) show corresponding sentences that contain the named mention `John', the nominal mention `the boy' and the pronominal mention `he'. This distinction between names on the one hand and pronominal or nominal mentions on the other can also be applied to the broad semantic set of named entities from the legal domain, see (\ref{lname}--\ref{lpronoun}). Thus, (\ref{name}, \ref{lname}) contain actual named entities.

\begin{exe}
\ex John is 8 years old.\label{name}
\ex The boy is 8 years old.\label{noun}
\ex He is 8 years old.\label{pronoun}
\ex The BGB regulates the legal relations between private persons.\label{lname}
\ex The law regulates the legal relations [\dots].\label{lnoun}
\ex It regulates the legal relations [\dots].\label{lpronoun}
\end{exe}

\paragraph{Legal Entity} Basically, legal entities are either designations or references. A designation (or name) is the title of a legal document. In law texts, the title is strictly standardised and consists of a long title, short title and an abbreviation \cite[margin nos 321 et seqq.]{handbuch2008}. The title of the Act on the Federal Constitutional Court is: `Gesetz {\"u}ber das Bundesverfassungsgericht (Bundesverfassungsgerichtsgesetz -- BVerfGG)', where `Gesetz {\"u}ber das Bundesverfassungsgericht' is the long title, `Bundesverfassungsgerichtsgesetz' is the short title, and `BVerfGG' is the abbreviation. A reference to a legal norm is also fixed with rules for short and full references \cite[margin nos 168 et seqq.]{handbuch2008}. Designations or references of binding individual acts such as regulations or contracts, however, are not uniformly defined.

\paragraph{Personal Data} A fundamental characteristic of the published decisions, that are the basis of our dataset, is that all personal information have been anonymised for privacy reasons. This affects the classes \emph{person}, \emph{location} and \emph{organization}. Depending on the respective federal court, different rules were used for this anonymisation process. Named entities were replaced by letters or abbreviated (\ref{anon1}), sometimes ellipsis were used (\ref{anon2}, \ref{anon3}). Some anonymised \emph{locations} are mentioned with terms such as ``street'', ``place'', ``avenue'', etc.~that are part of this named entity (\ref{anon3}).

\begin{exe}
\ex Fernsehmoderator \colorbox{Cyan!50!white}{G. \colorbox{white}{\textbf{PER}}}\label{anon1}
\trans `television presenter G.'
\ex Firma \colorbox{Periwinkle!50}{X\dots{} \colorbox{white}{\textbf{UN}}}\label{anon2}
\trans `company X\dots'
\ex in der \colorbox{GreenYellow}{A-Stra{\ss}e \colorbox{white}{\textbf{STR}}}
in \colorbox{LimeGreen}{\dots{} \colorbox{white}{\textbf{ST}}}\label{anon3}
\trans `in the A-Street in \dots{}'
\end{exe}

\subsection{Semantic Classes}

We defined 19 fine-grained semantic classes. The (proto)typical classes are \emph{person}, \emph{location} and \emph{organization}. In addition, we defined more specific semantic classes for the legal domain. These are the coarse-grained classes \emph{legal norm}, \emph{case-by-case regulation}, \emph{court decision} and \emph{legal literature}. The classes \emph{legal norm} and \emph{case-by-case regulation} include designations and references, while \emph{court decision} and \emph{legal literature} include only references.

In the process of developing the typology and annotation guidelines, the fine-grained classes \emph{continent} \texttt{KONT} (which belongs to \emph{location}), \emph{university} \texttt{UNI}, \emph{institute} \texttt{IS} and \emph{museum} \texttt{MUS} (which belonged to \emph{organization}) were eliminated due their low frequency in the corpus (less than 50 occurrences). This is why \emph{university}, \emph{institute} and \emph{museum} were subsumed under the fine-grained class  \emph{organization}. \emph{Continent} was integrated into \emph{landscape}.

The specification of the 19 fine-grained classes was motivated by the need for distinguishing entities in the legal domain. A first distinction was made between standards and binding acts. Standards, which belong to \emph{legal norm}, are legal rules adopted by a legislative body in a legislative process. We can distinguish further between \emph{law}, \emph{ordinance} (German national standards) and \emph{European legal norm}. Binding acts (circulars, administrative acts, contracts, administrative regulations, directives, etc.) belong to the category of \emph{case-by-case-regulation}. It includes \emph{regulation} (arrangements or instructions on subjects) and \emph{contract} (agreements between subjects). In addition, \emph{court decision} and \emph{legal literature}, which are important in the decision making process, were put into their own categories.

Within \emph{person}, we distinguish between \emph{judge} and \emph{lawyer}, key roles mentioned frequently in the decisions. Locations are categorised in terms of their size in \emph{country}, \emph{city} and \emph{street}. Organizations are divided based on their role in the process, into public or social \emph{organization}, state \emph{institution}, (private) economic \emph{company}, mostly as a legal entity, and \emph{court} as an organ of jurisprudence.

\paragraph{Person}

The coarse-grained class \emph{person} \texttt{PER} contains the fine-grained classes \emph{judge} \texttt{RR}, \emph{lawyer} \texttt{AN} and \emph{person} \texttt{PER} (such as accused, plaintiff, defendant, witness, appraiser, expert, etc.), who are involved in a court process and mentioned in a decision. In example (\ref{per}), the same surname occurs twice in a sentence, one as \emph{judge} and one as \emph{person}.

\begin{exe}
\ex Zwar ist \colorbox{Aquamarine!50}{Paul Kirchhof \colorbox{white}{\textbf{RR}}} mit dem Vize\-pr{\"a}\-si\-den\-ten \colorbox{Cyan!50}{Kirchhof \colorbox{white}{\textbf{PER}}} als dessen Bruder in der Seitenlinie im zweiten Grade verwandt\dots{}\label{per}
\trans `Although Paul Kirchhof is related to the Vice President Kirchhof as his brother in the second-degree sidelines\dots'
\end{exe}

\paragraph{Location}

The coarse-grained class \emph{location} \texttt{LOC} contains names of topographic objects, divided into \emph{country} \texttt{LD}, \emph{city} \texttt{ST}, \emph{street} \texttt{STR} and \emph{landscape} \texttt{LDS}. \emph{Country} (\ref{ld}) includes countries, states or city-states and \emph{city} (\ref{st}) includes to cities, villages or communities. \emph{Street} (\ref{str}) refers to avenues, squares, municipalities, attractions etc., i.\,e., named entities within a city or a village. \emph{Landscape} (\ref{lds}) includes continents, lakes, rivers and other geographical objects.

\begin{exe}
\ex \dots{} hat bislang nur das Land \colorbox{ForestGreen!50}{Mecklenburg-Vor-\colorbox{ForestGreen!50}{\textcolor{ForestGreen!50}{I}}} \colorbox{ForestGreen!50!white}{pommern \colorbox{white}{\textbf{LD}}} Gebrauch gemacht.\label{ld}
\trans `So far, only the state of Mecklenburg-Vorpommern has made use of it.'
\ex Dem Haftbefehl liegt eine Entscheidung des Berufungsgerichts in \colorbox{LimeGreen}{Bukarest \colorbox{white}{\textbf{ST}}} vom 18. Februar 2016 zugrunde \dots\label{st}
\trans `The arrest warrant is based on a decision of the Appeal Court in Bucharest of 18 February 2016 \dots'
\ex Zwar legt der Bezug auf die Grenz\-wert\-{\"u}b\-er\-schreitung 2015 insbesondere in der \colorbox{GreenYellow}{Cornelius-\colorbox{GreenYellow}{\textcolor{GreenYellow}{I}}} \colorbox{GreenYellow}{stra{\ss}e \colorbox{white}{\textbf{STR}}}  \dots\label{str}
\trans `Admittedly, the reference to the exceedance of the 2015 threshold applies in particular to Corneliusstra{\ss}e \dots'
\ex \dots{} aus der Region um den Fluss \colorbox{SpringGreen!70!white}{Main \colorbox{white}{\textbf{LDS}}} stammen bzw. dort angeboten werden \dots\label{lds}
\trans `\dots{} come from the region around the river Main or are offered there\dots'
\end{exe}

\paragraph{Organization}

The coarse-grained class \emph{organization} \texttt{ORG} is divided into public/social, state and economic institutions. Social and public institutions such as parties, associations, centres, communities, unions, educational institutions or research institutions are grouped into the fine-grained class \emph{organization} \texttt{ORG} (\ref{org}).  \emph{Institution} \texttt{INN} (\ref{inn}) contain public administrations, including federal and state ministries and the constitutional bodies of the Federal Republic of Germany at the federal and state level, i.\,e., the Federal Government, the Federal Council, the Bundestag, the state parliaments and governments. \emph{Company} \texttt{UN} (\ref{un}) includes commercial legal entities.

\begin{exe}
\ex Der \colorbox{Thistle!50!white}{FC Bayern M{\"u}nchen \colorbox{white}{\textbf{ORG}}} schloss den Be\-schwer\-def{\"u}hrer \dots{} aus dem Verein aus \dots\label{org} 
\trans `Bayern Munich closed the complainant \dots{} from the club \dots{}.'
\ex Die \colorbox{Orchid!50!white}{Landesregierung Rheinland-Pfalz \colorbox{white}{\textbf{INN}}} hat von einer Stellungnahme abgesehen.\label{inn} 
\trans `The state government of Rhineland-Palatinate refrained from commenting.'
\ex \dots{} eingef{\"u}hrte Smartphone-Mo\-d\-ell\-reihe des US-amerikanischen Unternehmens \colorbox{Periwinkle!50}{Apple \colorbox{white}{\textbf{\textbf{UN}}}} \dots\label{un}
\trans `\dots{} introduced smartphone model series of the US company Apple \dots'
\end{exe}

Court designations play a central role in decisions, which is why they are collected in their own class \emph{court} \texttt{GRT}. These are designations of federal, supreme, provincial and local courts. The designations of the courts at the country level are composed of the names of the ordinary jurisdiction and their location (\ref{grt1}). Furthermore, brands are often discussed in decisions of the Federal Patent Court. They are subsumed under \emph{brand} \texttt{MRK}, which can be contextual and semantically ambiguous, such as `Becker' from (\ref{mrk}), which has evolved from a personal name.

\begin{exe}
\ex Diesen Anspruch hat das \colorbox{RubineRed!50}{LSG Mecklenburg-\colorbox{RubineRed!50}{\textcolor{RubineRed!50}{I}}} \colorbox{RubineRed!50!white}{Vorpommern \colorbox{white}{\textbf{GRT}}} mit Urteil vom 22.2.2017 verneint \dots\label{grt1}
\trans `This claim was rejected by the LSG Mecklenburg-Vorpommern by judgment of 22.2.2017 \dots'
\ex Vorliegend stehen sich die Widerspruchsmarke \colorbox{Salmon!70}{Becker Mining \colorbox{white}{\textbf{MRK}}} und die angegriffene Marke \colorbox{Salmon!70}{Becker \colorbox{white}{\textbf{MRK}}} gegen{\"u}ber.\label{mrk}
\trans `In the present case, the opposing brand Becker Mining and the challenged brand Becker face each other.'
\end{exe}

\paragraph{Legal Norms}

Norms are divided according to their legal status into the fine-grained classes of \emph{law} \texttt{GS}, \emph{ordinance} \texttt{VO} and \emph{European legal norm} \texttt{EUN}. \emph{Law} is composed of the standards adopted and designated by the legislature (Bundestag, Bundesrat, Landtag). \emph{Ordinance} includes standards adopted by a federal or provincial government or by a ministry. \emph{European legal norm} includes norms of European primary or secondary legislation, European organizations and other conventions and agreements.

Example (\ref{gs1}) includes a reference to the `Part-Time and Limited Term Employment Act' and the designation 'Basic Law'. The complex reference consists of the reference to the particular section of the law, its name and abbreviation (in brackets), date of issue, the reference in parenthesis and the details of the most recent change. Cases such as this one are a full reference. Example (\ref{vo1}), on the other hand, shows a short reference consisting of information on the corresponding section of the law and the abbreviated name of the statutory order.

\begin{exe}
\ex \label{gs1} \dots{}\,\colorbox{RedOrange!70}{\S{}\,14\,Absatz\,2\,Satz\,2 des Gesetzes {\"u}ber Teil-\,\colorbox{RedOrange!70}{\textcolor{RedOrange!70}{I}}} 
\colorbox{RedOrange!70}{zeitarbeit und {I}\textcolor{RedOrange!70}{I}befristete Arbeitsvertr{\"a}ge \textcolor{RedOrange!70}{I}(Tz-\colorbox{RedOrange!70}{\textcolor{RedOrange!70}{I}}}
\colorbox{RedOrange!70}{BfG) vom 21. Dezember 2000 (Bundesgesetz- \colorbox{RedOrange!70}{\textcolor{RedOrange!70}{I}}} \colorbox{RedOrange!70}{blatt Seite 1966), zuletzt ge{\"a}ndert durch Gesetz\colorbox{RedOrange!70}{\textcolor{RedOrange!70}{I}}} 
\colorbox{RedOrange!70}{vom\textcolor{RedOrange!70}{I}\textcolor{RedOrange!70}{I}20.\textcolor{RedOrange!70}{I}Dezember\textcolor{RedOrange!70}{I}2011 (Bundesgesetzblatt I\colorbox{RedOrange!70}{\textcolor{RedOrange!70}{I}}} \colorbox{RedOrange!70}{Seite 2854\,) \textcolor{RedOrange!70}{I} \colorbox{white}{\textbf{GS}}}, ist nach Ma{\ss}gabe der Gr{\"u}nde mit dem \colorbox{RedOrange!70}{Grundgesetz \colorbox{white}{\textbf{GS}}} vereinbar. 
\trans `\dots{} section 14 paragraph 2 sentence 2 of the Law on Part-Time and Limited Term Employment Act (TzBfG) of 21 December 2000 (Federal Law Gazette I, page 1966), as last amended by the Law of 20 December 2011 (Federal Law Gazette I, page 2854), shall be published in accordance with the reasons compatible with the Basic Law.'
\ex \label{vo1}
\trans \dots{} Neuregelung in \colorbox{BurntOrange!50}{\S{} 35 Abs. 6 StVO \colorbox{white}{\textbf{VO}}}\dots
\trans `\dots{} new regulation in sec. 35 para. 6 StVO\dots'
\end{exe}

\paragraph{Case-by-case Regulation}

The class \emph{case-by-case regulation} \texttt{REG} contains individual binding acts. These include \emph{regulation} \texttt{VS} and \emph{contract} \texttt{VT}. \emph{Regulation} is an internal order or instruction from a superordinate authority to a subordinate, regulating their activities. In addition to administrative regulations, these include guidelines, circulars and decrees. In contrast to \emph{legal norm}, these rules have no direct effect on the citizen. The class \emph{contract} includes public contracts, international treaties and collective agreements. Some designations and references from these classes are similar to \emph{legal norm} (\ref{vs1}, \ref{vt1}).

\begin{exe}
\ex \label{vs1} \dots{} insbesondere durch die \colorbox{Peach!70}{ Richtlinien zur Be-\colorbox{Peach!70}{\textcolor{Peach!70}{I}}}
\colorbox{Peach!70}{wertung des Grundverm{\"o}gens\,--BewRGr--\,vom \colorbox{Peach!70}{\textcolor{Peach!70}{I}}} \colorbox{Peach!70}{19. September 1966 (BStBl I, S. 890) \colorbox{white}{\textbf{VS}}}.
\trans `\dots{} in particular by the Guidelines for the Valuation of Real Estate -- BewRGr -- of 19 September 1966 (BStBl I, p. 890).'
\ex \label{vt1}
\dots{}\,fand der \colorbox{Goldenrod!70}{Manteltarifvertrag f{\"u}r die Besch{\"a}f-\colorbox{Goldenrod!70}{\textcolor{Goldenrod!70}{I}}} \colorbox{Goldenrod!70}{tigten der Mitglieder der TGAOK\,\colorbox{white}{\textbf{VT}}}\,(\colorbox{Goldenrod!70}{BAT/\colorbox{Goldenrod!70}{\textcolor{Goldenrod!70}{I}}} \colorbox{Goldenrod!70}{AOK-Neu \colorbox{white}{\textbf{VT}}}) vom 7. August 2003 Anwendung.
\trans `\dots{} the Collective Agreement for the Employees of Members of TGAOK (BAT/AOK-New) was applied of 7 August 2003 \dots'
\end{exe}

\paragraph{Court Decision}

The class \emph{court decision} \texttt{RS} includes references to decisions. It does not have any subclasses, the coarsed and fine-grained versions are identical. In \emph{court decision}, the name of the official decision-making collection, the volume and the numbered article are cited. Often mentioned are also the court, if necessary the decision type, date and file number. Example (\ref{rs1}) cites decisions of the Federal Constitutional Court (BVerfG) and the Federal Social Court (BSG). Decisions of the BVerfG are referenced with regard to pages, while decisions of the BSG are sorted according to paragraphs, numbers and marginal numbers.

\paragraph{Legal Literature}

\emph{Legal literature} \texttt{LIT} also contains references, but they refer to legal commentaries, legislative material, legal textbooks and monographs. The commentary in example (\ref{rs1}) includes the details of an author's and/or publisher's name, the name of a legal norm, a paragraph and a paragraph number. Multiple authors are separated by a slash. Textbooks and monographs are cited as usual (author's name, title, edition, year of publication, page number). References of legislative materials consist of a title and reference marked with numbers.

\begin{exe}
\ex \label{rs1} 
\dots{} vgl zB \colorbox{Dandelion!70}{BVerfGE 62, 1, 45 \colorbox{white}{\textbf{RS}}}; \colorbox{Dandelion!70}{BVerfGE\colorbox{Dandelion!70}{\textcolor{Dandelion!70}{I}}} \colorbox{Dandelion!70}{119,\,96,\,179 \colorbox{white}{\textbf{RS}}}; \colorbox{Dandelion!70}{BSG SozR 4--2500\,\S{}\,62 Nr\colorbox{Dandelion!70}{\textcolor{Dandelion!70}{I}}}
\colorbox{Dandelion!70}{8\,RdNr\,20\,f\,\colorbox{white}{\textbf{RS}}};
\colorbox{Tan!60!Bittersweet!70!white}{Hauck/Wiegand,\,KrV 2016,\colorbox{Tan!60!Bittersweet!70!white}{\textcolor{Tan!60!Bittersweet!70!white}{I}}}
\colorbox{Tan!60!Bittersweet!70!white}{1, 4 \colorbox{white}{\textbf{LIT}}} \dots
\trans `\dots{} cf. i.e. BVerfGE 62, 1, 45; BVerfGE 119, 96, 179; BSG SozR 4--2500\,\S{}\,62 Nr 8 RdNr 20 f; Hauck/Wiegand, KrV 2016, 1, 4 \dots'
\end{exe}

\section{Description of the Dataset}
\label{sec:dataset}

The dataset\footnote{\url{https://github.com/elenanereiss/Legal-Entity-Recognition}}, which also includes annotation guidelines, is freely available under a CC-BY 4.0 license.\footnote{\url{https://creativecommons.org/licenses/by/4.0/deed.en}} The named entity annotations adhere to the CoNLL-2002 format \cite{sangerik}, while time expressions were annotated using TimeML \cite{pustejovsky2003timeml}.

\subsection{Original Source Documents}

Legal documents are a rather heterogeneous class, which also manifests in their linguistic properties, including the use of named entities and references. Their type and frequency varies significantly, depending on the text type. Texts belonging to specific text type, which are to be selected for inclusion in a corpus must contain enough different named entities and references and they need to be freely available. When comparing legal documents such as laws, court decisions or administrative regulations, decisions are the best option. In laws and administrative regulations, the frequencies of \emph{person}, \emph{location} and \emph{organization} are not high enough for NER experiments. Court decisions, on the other hand, include \emph{person}, \emph{location}, \emph{organization}, references to \emph{law}, other \emph{decision} and \emph{regulation}.

%\subsubsection{Extracted Data}

Court decisions from 2017 and 2018 were selected for the dataset, published online by the Federal Ministry of Justice and Consumer Protection.\footnote{\url{https://www.rechtsprechung-im-internet.de}} The documents originate from seven federal courts: Federal Labour Court (BAG), Federal Fiscal Court (BFH), Federal Court of Justice (BGH), Federal Patent Court (BPatG), Federal Social Court (BSG), Federal Constitutional Court (BVerfG) and Federal Administrative Court (BVerwG). 

From the table of contents\footnote{\url{http://www.rechtsprechung-im-internet.de/rii-toc.xml}}, 107 documents from each court were selected (see Table~\ref{court_distrib}). The data was collected from the XML documents, i.\,e., it was extracted from the XML elements \texttt{Mit\-wir\-kung}, \texttt{Ti\-tel\-zei\-le}, \texttt{Le\-it\-satz}, \texttt{Te\-n\-or}, \texttt{Tatbestand}, \texttt{Entscheidungsgr{\"u}nde}, \texttt{Gr{\"u}nden}, \texttt{abweichende Meinung}, and \texttt{sonstiger Titel}. The metadata at the beginning of the documents (name of court, date of decision, file number, European Case Law Identifier, document type, laws) and those that belonged to previous legal proceedings was deleted. Paragraph numbers were removed. The extracted data was split into sentences, tokenised using SoMaJo\footnote{\url{https://github.com/tsproisl/SoMaJo}} \cite{somajo} and manually annotated in WebAnno\footnote{\url{https://webanno.github.io/webanno/}} \cite{webanno3}.

The annotated documents are available in CoNNL-2002. The information originally represented by and through the XML markup was lost in the conversion process. We decided to use CoNNL-2002 because our primary focus was on the NER task and experiments. CoNNL is one of the best practice formats for NER datasets. All relevant tools support CoNNL, including WebAnno for manual annotation. Nevertheless, it is possible, of course, to re-insert the annotated information back into the XML documents.

\subsection{Annotation of Named Entities}

%\subsubsection{Size of Dataset and Subdatasets}

The dataset consists of 66,723 sentences with 2,157,048 tokens (incl.~punctuation), see Table~\ref{corpus_size}. The sizes of the seven court-specific datasets varies between 5,858 and 12,791 sentences, and 177,835 to 404,041 tokens. The distribution of annotations on a per-token basis corresponds to approx.~19--23\,\%. The Federal Patent Court (BPatG) dataset contains the lowest number of annotated entities (10.41\,\%).

\input{tables/size.tex}

%\subsubsection{Distribution of Entities}

The dataset includes two different versions of annotations, one with a set of 19 fine-grained semantic classes and another one with a set of 7 coarse-grained classes (Table~\ref{stat_fcg}). There are 53,632 annotated entities in total, the majority of which (74.34\,\%) are legal entities, the others are \emph{person}, \emph{location} and \emph{organization} (25.66\,\%). Overall, the most frequent entities are \emph{law} \texttt{GS} (34.53\,\%) and \emph{court decision} \texttt{RS} (23.46\,\%). The other legal classes (\emph{ordinance} \texttt{VO}, \emph{European legal norm} \texttt{EUN}, \emph{regulation} \texttt{VS}, \emph{contract} \texttt{VT}, and \emph{legal literature} \texttt{LIT}) are much less frequent (1--6\,\% each). Even less frequent (less than 1\,\%) are \emph{lawyer} \texttt{AN}, \emph{street} \texttt{STR}, \emph{landscape} \texttt{LDS}, and \emph{brand} \texttt{MRK}.

\input{tables/stat_fcg.tex}

The classes \emph{person}, \emph{lawyer} and \emph{company} are heavily affected by the anonymisation process (80\,\%, 95\,\% and 70\,\% respectively). More than half of \emph{city} and \emph{street}, about 55\,\%, have also been modified. \emph{Landscape} and \emph{organization} are affected as well, with 40\,\% and 15\,\% of the occurrences edited accordingly. However, anonymisation is typically not applied to \emph{judge}, \emph{country}, \emph{institution} and \emph{court} (1--5\,\%). 

%\subsubsection{Inter-Annotator Agreement}

The dataset was originally annotated by the first author. To evaluate and potentially improve the quality of the annotations, part of the dataset was annotated by a second linguist (using the annotation guidelines specifically prepared for its construction). We selected a small part that could be annotated in approx.~two weeks. For the sentence extraction we paid special attention to the anonymised mentions of \emph{person}, \emph{location} or \emph{organization} entities, because these are usually explained at their first mention. The resulting sample consisted of 2005 sentences with a broad variety of different entities (3\,\% of all sentences from each federal court). The agreement between the two annotators was measured using Kappa on a token basis. All class labels were taken into account in accordance with the IOB2 scheme \cite{sang1999veenstra}. The inter-annotator agreement is 0.89, i.\,e., there is mostly very good agreement between the two annotators. Differences were in the identification of \emph{court decision} and \emph{legal literature}. Some unusual references of \emph{court decision} (consisting only of decision type, court, date, file number) were not annotated such as `Urteil des Landgerichts Darmstadt vom 16.~April 2014 -- 7 S 8/13 --'. Apart from missing \emph{legal literature} annotations, author names and law designations were annotated according to their categories (i.\,e., `Schoch, in: Schoch/Schneider/Bier, VwGO \S{} 123 Rn.~35', `Bekanntmachung des BMG gem{\"a\ss} \S\S{} 295 und 301 SGB V zur Anwendung des OPS vom 21.10.2010').

The second annotator had difficulties annotating the class \emph{law}, not all instances were identified (`\S{} 272 Abs.~1a und 1b HGB', `\S{} 3c Abs.~2 Satz 1 EStG'), others only partially (`\S{} 716 in Verbindung mit' in `\S{} 716 in Verbindung mit \S\S{} 321 , 711 ZPO'). Some titles of \emph{contract} were not recognised and annotated (`BAT', `TV-L', `TV{\"U}-L{\"a}nder' etc.).

This evaluation has revealed deficiencies in the annotation guidelines, especially regarding \emph{court decision} and \emph{legal literature} as well as non-entities. It would also be helpful for the identification and classification to list well-known sources of \emph{law}, \emph{court decision}, \emph{legal literature} etc.

\subsection{Annotation of Time Expressions}

All court decisions were annotated automatically for time expressions using a customised version of HeidelTime \cite{StroetgenGertz2013:LREjournal}, which was adapted to the legal domain \cite{Weissenhorn2018}. This version of Heideltime achieves an F$_1$ value of 89.1 for partial identification and normalization. It recognizes four \texttt{TIMEX3}-types of time expressions \cite{verhagen-etal-2010-semeval}: \texttt{DATE}, \texttt{DURATION}, \texttt{SET}, \texttt{TIME}. \texttt{DATE} describe a calendar date (`23.~July 1994', `November 2019', `winter 2001' etc). It also includes expressions such as `present', `former' or `future'. \texttt{DURATION} describes time periods such as `two hours' or `six years'. \texttt{SET} describes a set of times/periods (`every day', `twice a week'). \texttt{TIME} describes a time expression (`13:12', `tomorrow afternoon'). Expressions with a granularity less than 24 hours are of type \texttt{TIME}, all others are of type \texttt{DATE}. The distribution of \texttt{TIMEX3} types in the legal dataset is shown in Table~\ref{time} with a total number of 35,119 time expressions, approx.~94\.\% of which are of type \texttt{DATE}.

\begin{exe}
\ex \dots vgl. BGH, Beschluss vom \textcolor{blue}{$<$TIMEX3} \textcolor{red}{tid}=\textcolor{Purple}{"t14"} \textcolor{red}{type}=\textcolor{Purple}{"DATE"} \textcolor{red}{value}=\textcolor{Purple}{''1999-02-03''}\textcolor{blue}{$>$}3.~Februar 1999\textcolor{blue}{$<$/TIMEX3$>$} -- 5 StR 705/98, juris Rn.~2 \dots
\end{exe}

\input{tables/time.tex}

\section{Evaluation}
\label{sec:evaluation}

The dataset was thoroughly evaluated, see \newcite{leitner2019fine} for more details. As state of the art models, Conditional Random Fields (CRFs) and bidirectional Long-Short Term Memory Networks (BiLSTMs) were tested with the two variants of annotation. For CRFs, these are: CRF-F (with features), CRF-FG (with features and gazetteers), CRF-FGL (with features, gazetteers and lookup). For BiLSTM, we used models with pre-trained word embeddings \cite{reimers2014germeval}: BiLSTM-CRF \cite{huang2015bidirectional}, BiLSTM-CRF+ with character embeddings from BiLSTM \cite{lample2016neural}, and BiLSTM-CNN-CRF with character embeddings from CNN \cite{ma2016end}. To evaluate the performance we used stratified 10-fold cross-validation. As expected, BiLSTMs perform best (see Table~\ref{eval}). The F$_1$ score for the fine-grained classification reaches 95.46 and 95.95 for the coarse-grained one. CRFs reach up to 93.23 F$_1$ for the fine-grained classes and 93.22 F$_1$ for the coarse-grained ones. Both models perform best for \emph{judge}, \emph{court} and \emph{law}. 

\input{tables/eval.tex}

\section{Conclusions and Future Work}
\label{sec:conclusion}

We describe a dataset that consists of German legal documents. For the annotation, we specified a typology of characteristic semantic categories that are relevant for court decisions (i.\,e., \emph{court}, \emph{institution}, \emph{law}, \emph{court decision}, and \emph{legal literature}) with corresponding annotation guidelines. A functional service based on the work described in this paper will be made available through the European Language Grid \cite{rehm2020m}.

In terms of future work, we will look into approaches for extending and further optimizing the dataset. We will also perform additional experiments with more recent state of the art approaches (i.\,e., with language models); preliminary experiments using BERT failed to yield an improvement. We also plan to replicate the dataset in one or more other languages, such as English, Spanish, or Dutch, to cover at least one more of the relevant languages in the Lynx project. We also plan to produce an XML version of the dataset that also includes the original XML annotations.

\section*{Acknowledgements}

This work has been partially funded by the project Lynx, which has received funding from the EU's Horizon 2020 research and innovation programme under grant agreement no.~780602, see \url{http://www.lynx-project.eu}.

\section{References}
\label{main:ref}

\bibliographystyle{./lrec}
\bibliography{./lrec2020}

%\section{Language Resource References}
%\label{lr:ref}
%\bibliographystylelanguageresource{lrec}
%\bibliographylanguageresource{lrec2020W-xample}

\end{document}

%% file: tables/size.tex
\begin{table}[ht]
\centering
\begin{tabular}{lcrrr}
\toprule
\multicolumn{1}{c}{\textbf{}} & 
\multicolumn{1}{c}{\textbf{Docu-}} & 
\multicolumn{1}{c}{\textbf{}} & 
\multicolumn{1}{c}{\textbf{Sent-}} & \multicolumn{1}{c}{\textbf{Annotated}} \\
\multicolumn{1}{c}{\textbf{Court}}                & \multicolumn{1}{c}{\textbf{ments}} & \multicolumn{1}{c}{\textbf{Tokens}}                & \multicolumn{1}{c}{\textbf{ences}}                   & \multicolumn{1}{c}{\textbf{tokens}} \\ \midrule
BAG                                   & 107                                    & 343,065                              & 12,791                                  & 19.23\%                                   \\ \midrule
BFH                                   & 107                                    & 276,233                              & 8,522                                   & 22.43\%                                   \\ \midrule
BGH                                   & 108                                    & 177,835                              & 5,858                                   & 19.23\%                                   \\ \midrule
BPatG                                 & 107                                    & 404,041                              & 12,016                                  & 10.41\%                                   \\ \midrule
BSG                                   & 107                                    & 302,161                              & 8,083                                   & 22.76\%                                   \\ \midrule
BVerfG                                & 107                                    & 305,889                              & 9,237                                   & 22.09\%                                   \\ \midrule
BVerwG                                & 107                                    & 347,824                              & 10,216                                  & 20.84\%                                   \\ \midrule
\textbf{Total}                        & \textbf{750}                           & \textbf{2,157,048}                   & \textbf{66,723}                         & \textbf{19.15\%}                          \\ \bottomrule
\end{tabular}
\caption{Dataset size (tokens, sentences, annotated tokens)}
\label{court_distrib}
\label{corpus_size}
\end{table}

%% file: tables/stat_fcg.tex
\begin{table}[h]
\centering
\begin{tabular}{lrllrr} \toprule
\multicolumn{4}{c}{\textbf{Classes}} & \multicolumn{1}{c}{\textbf{\#}} & \multicolumn{1}{c}{\textbf{\%}} \\ \midrule
f & 1         & \texttt{PER}        & Person                     & 1,747                           & 3.26                             \\ 
f & 2         & \texttt{RR}         & Judge                      & 1,519                           & 2.83                             \\ 
f & 3         & \texttt{AN}         & Lawyer                     & 111                              & 0.21                             \\ 
\textbf{c} & \textbf{1}          & \textbf{\texttt{PER}}         & \textbf{Person}                    & \textbf{3,377}                           & \textbf{6.30}                             \\ \midrule 
f & 4         & \texttt{LD}         & Country                    & 1,429                           & 2.66                             \\ 
f & 5         & \texttt{ST}         & City                       & 705                              & 1.31                             \\ 
f & 6         & \texttt{STR}        & Street                     & 136                              & 0.25                             \\ 
f & 7         & \texttt{LDS}        & Landscape                  & 198                              & 0.37                             \\ 
\textbf{c} & \textbf{2}          & \textbf{\texttt{LOC}}          & \textbf{Location}                  & \textbf{2,468}                           & \textbf{4.60}                             \\ \midrule
f & 8         & \texttt{ORG}        & Organization               & 1,166                           & 2.17                             \\ 
f & 9         & \texttt{UN}         & Company                    & 1,058                           & 1.97                             \\ 
f & 10        & \texttt{INN}        & Institution                & 2,196                           & 4.09                             \\ 
f & 11        & \texttt{GRT}        & Court                      & 3,212                           & 5.99                             \\ 
f & 12        & \texttt{MRK}        & Brand                      & 283                              & 0.53                             \\ 
\textbf{c} & \textbf{3}          & \textbf{\texttt{ORG}}          & \textbf{Organization}              & \textbf{7,915}                           & \textbf{14.76}                            \\ \midrule
f & 13        & \texttt{GS}         & Law                        & 18,520                          & 34.53                            \\ 
f & 14        & \texttt{VO}         & Ordinance                  & 797                              & 1.49                             \\ 
f & 15        & \texttt{EUN}        & EU legal norm        & 1,499                           & 2.79                             \\ 
\textbf{c} & \textbf{4}          & \textbf{\texttt{NRM}}          & \textbf{Legal norm}                & \textbf{20,816}                          & \textbf{38.81}                            \\
\midrule
f & 16        & \texttt{VS}         & Regulation                 & 607                              & 1.13                             \\ 
f & 17        & \texttt{VT}         & Contract                   & 2,863                           & 5.34                             \\ 
\textbf{c} & \textbf{5}          & \textbf{\texttt{REG}}          & \textbf{Case-by-c. regul.}              & \textbf{3,470}                           & \textbf{6.47}                             \\ \midrule
f & 18      &        &   &   &   \\ 
\textbf{c} & \textbf{6}      & \textbf{\texttt{RS}}         & \textbf{Court decision}             & \textbf{12,580}                          & \textbf{23.46}                            \\ \midrule
f & 19        &         &           &                         &              \\ 
\textbf{c} & \textbf{7}        & \textbf{\texttt{LIT}}        & \textbf{Legal literature}           & \textbf{3,006}                           & \textbf{5.60}                             \\ \midrule
\multicolumn{4}{l}{\textbf{Total}}                & \textbf{53,632}                 & \textbf{100}                     \\ \bottomrule
\end{tabular}
\caption{Distribution of fine-grained (f) and coarse-grained (c) classes in the dataset}
\label{stat_fcg}
\end{table}

%% file: tables/time.tex
\begin{table}[h]
\centering
\begin{tabular}{lrrrr}
\toprule
\multicolumn{1}{c}{\textbf{}} &\texttt{DATE}           & \texttt{DURATION}      &\texttt{SET}          &\texttt{TIME}         \\ \midrule
BAG                                  & 6,463           & 491           & 99           & 34           \\ \midrule
BFH                                  & 6,156           & 189           & 37           & 9            \\ \midrule
BGH                                  & 2,819           & 254           & 7            & 22           \\ \midrule
BPatG                                & 4,576           & 84            & 4            & 12           \\ \midrule
BSG                                  & 4,634           & 215           & 64           & 14           \\ \midrule
BVerfG                               & 3,595           & 207           & 12           & 20           \\ \midrule
BVerwG                               & 4,879           & 178           & 36           & 9            \\ \midrule
\textbf{Total}                       & \textbf{33,122} & \textbf{1,618} & \textbf{259} & \textbf{120} \\ \bottomrule
\end{tabular}
\caption{Distribution of time expressions in the dataset}\label{time}
\end{table}

%% file: tables/eval.tex
\begin{table}[h]
\centering
\begin{tabular}{llll}
\toprule
\textbf{}               & \multicolumn{1}{c}{\textbf{Prec \%}} & \multicolumn{1}{c}{\textbf{Rec \%}} & \multicolumn{1}{c}{\textbf{F$_1$}} \\ \midrule
\multicolumn{4}{c}{\textit{Annotation with fine-grained semantic classes}} \\
\textbf{CRF-F}          & 94.28            & 91.85           & 93.05       \\
\textbf{CRF-FG}         & 94.31            & 91.96           & 93.12       \\
\textbf{CRF-FGL}        & 94.37            & 92.12           & \textbf{93.23}       \\ \midrule
\multicolumn{4}{c}{\textit{Annotation with coarse-grained semantic classes}} \\
\textbf{CRF-F}          & 94.17            & 92.07           & 93.11       \\
\textbf{CRF-FG}         & 94.26            & 92.20           & \textbf{93.22}       \\
\textbf{CRF-FGL}        & 94.22            & 92.25           & \textbf{93.22}       \\ \midrule
\multicolumn{4}{c}{\textit{Annotation with fine-grained semantic classes}} \\
\textbf{BiLSTM-CRF}     & 93.80            & 93.70           & 93.75       \\
\textbf{BiLSTM-CRF+}    & 95.36            & 95.57           & \textbf{95.46}       \\
\textbf{BiLSTM-CNN-CRF} & 95.34            & 95.58           & \textbf{95.46}       \\ \midrule
\multicolumn{4}{c}{\textit{Annotation with coarse-grained semantic classes}} \\
\textbf{BiLSTM-CRF}     & 94.86            & 94.49           & 94.68       \\
\textbf{BiLSTM-CRF+}    & 95.84            & 96.07           & \textbf{95.95}       \\
\textbf{BiLSTM-CNN-CRF} & 95.71            & 95.87           & 95.79       \\ \bottomrule
\end{tabular}
\caption{Precision, recall and F$_1$ values of the CRF and Bi\-LSTM models for the fine- and coarse-grained classes}
\label{eval}
\end{table}